\documentclass[letterpaper, 10 pt, conference]{ieeeconf}  

\IEEEoverridecommandlockouts                              

\usepackage{graphics} 
\usepackage{graphicx} 
\usepackage{algorithm}
\usepackage{algorithmic}
\usepackage{threeparttable}

\usepackage{newtxmath}

\usepackage{caption}
\usepackage{subcaption}

\title{\LARGE \bf
Keyframe Demonstration Seeded and Bayesian Optimized Policy Search
}

\author{\authorblockN{Onur Berk Töre}
\authorblockA{KUIS AI Center\\
Koc University\\
Istanbul,\\
Email: otore19@ku.edu.tr}
\and
\authorblockN{Farzin Negahbani}
\authorblockA{KUIS AI Center\\
Koc University\\
Istanbul,\\
Email: fnegahbani19@ku.edu.tr}
\and
\authorblockN{Barış Akgün}
\authorblockA{KUIS AI Center\\
Koc University\\
Istanbul,\\
Email: baakgun@ku.edu.tr}}

\usepackage{xcolor}

\newcommand{\mycomment}[1]{}

\begin{document}

\maketitle
\thispagestyle{empty}
\pagestyle{empty}

\begin{abstract}

This paper introduces a novel Learning from Demonstration framework to learn robotic skills with keyframe demonstrations using a Dynamic Bayesian Network (DBN) and a Bayesian Optimized Policy Search approach to improve the learned skills. DBN learns the robot motion, perceptual change in the object of interest (aka skill sub-goals) and the relation between them. The rewards are also learned from the perceptual part of the DBN. The policy search part is a semi-black box algorithm, which we call BO-PI$^2$. It utilizes the action-perception relation to focus the high-level exploration, uses Gaussian Processes to model the expected-return and performs Upper Confidence Bound type low-level exploration for sampling the rollouts.
BO-PI$^2$ is compared against a state-of-the-art method on three different skills in a real robot setting with expert and naive user demonstrations. The results show that our approach successfully focuses the exploration on the failed sub-goals and the addition of reward-predictive exploration outperforms the state-of-the-art approach on cumulative reward, skill success, and termination time metrics.
\end{abstract}

\section{INTRODUCTION} \label{sec:intro}





Learning from demonstration (LfD) and Reinforcement Learning (RL) are promising alternatives to direct programming to endow robots with skills. However, they have their own challenges. LfD suffers from distribution shift, noisy demonstrations, and amount of data due to limited teacher patience. RL suffers from sample complexity and reward formulation (not all skills have easy to define rewards and an expert is needed to define them even if they do). 

LfD seeded RL alleviates the sample complexity problem by learning an initial skill from demonstrations, which reduces the exploration effort by starting from a favorable point. Nevertheless, this approach still requires a non-trivial amount of trials, and the challenges of getting good demonstrations from users and reward engineering remain.



In this paper, we are interested in endowing manipulators with skills that have perceptual (sub-)goals. We introduce an LfD seeded RL approach to tackle the aforementioned challenges within this context. Our developed approach: 
\begin{itemize}
\item Works with keyframe demonstrations since they were found to be easier to use by non-roboticist users \cite{keyframe_based_learning_from_demonstrations}.
\item Learns a probabilistic skill model from keyframe demonstrations that jointly represents the end-effector motion and perceptual states of the object of interest. 
\item Uses the perceptual part of the model to learn rewards to remove the need for specifying them. 
\item Leverages the relationship between action and perception to focus the exploration to reduce number of trials. 
\item Performs policy search with Upper Confidence Bound (UCB) based exploration to update skill model parameters to further reduce the number of trials.
\end{itemize}

\begin{figure}
\centering
\includegraphics[width=0.77\columnwidth]{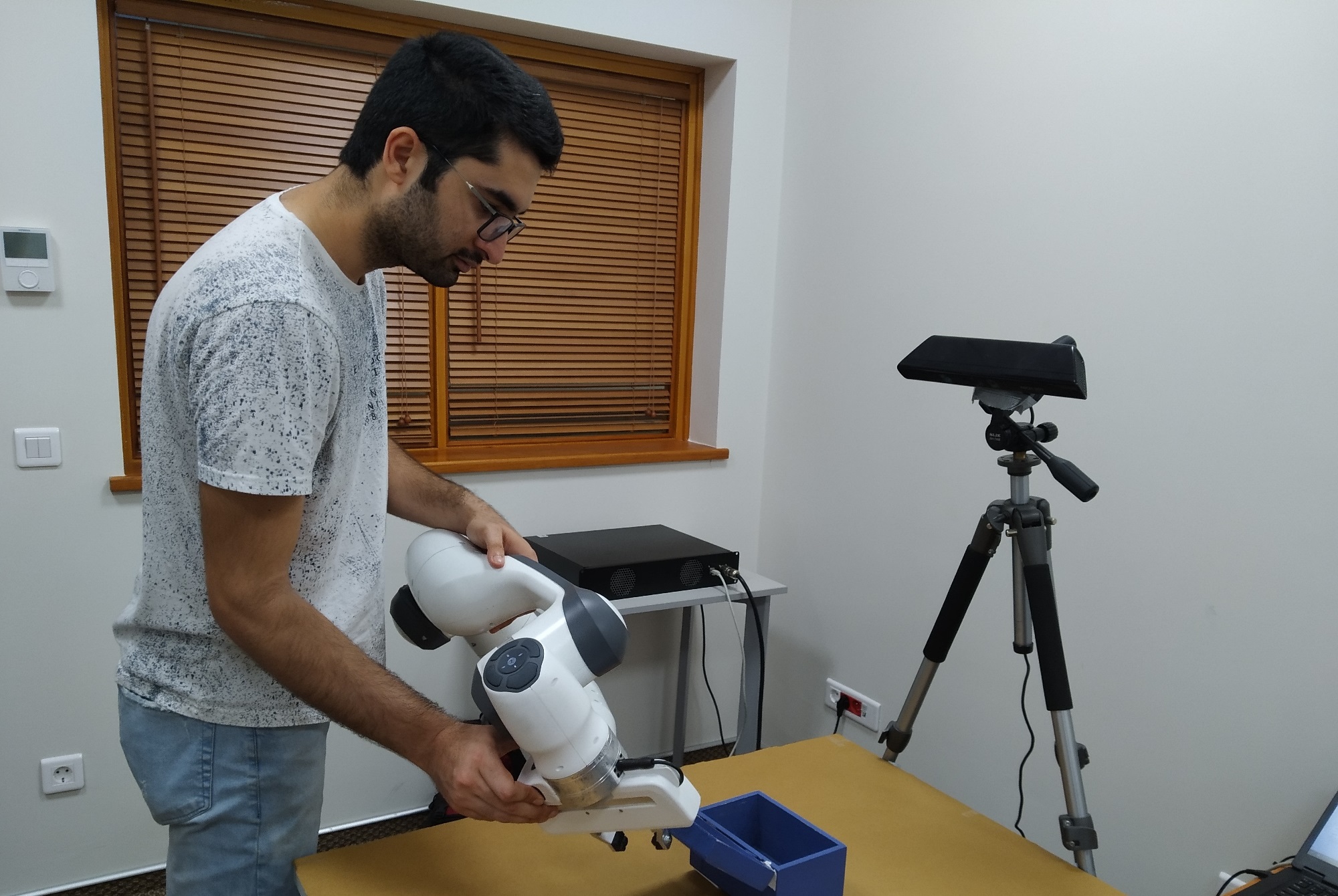}
\caption[Environment Setup]{Kinesthetic demonstration of the \textit{close the box} skill to a Franka Emika Panda arm. Microsoft Kinect V1 records Point Cloud data of the object of interest.}
\label{fig:setup}
\end{figure}



We develop a Dynamic Bayesian Network (DBN) as our probabilistic skill model. The DBN in our LfD framework is similar to a Hidden Markov Model (HMM), but it has two hidden states to represent the action and the perceptual (sub-)goals. The transition model of our DBN is composed of two tensors, modeling the probability of the next action/goal state with the current action and goal state, leading to learning the relationship between actions and sub-goals. The DBN model provides a connection between the robot motion and the skill sub-goals. We use this to select action state pairs to concentrate the RL effort by determining the part of the motion that does not reach the desired sub-goals.


We update the emission parameters of the action part, which can be interpreted as our policy, during RL with the reward generated by the goal part. For this, we use a novel Black-Box RL Policy Search (PS) method, called the Bayesian Optimized PI$^{2}$ (BO-PI$^{2}$), to update the policy parameters and use the reward learning ideas in \cite{reward_learning_from_very_few_demonstrations}.

We test our approach with real-world expert and naive user demonstrations involving three skills with the setup shown in Fig.\ref{fig:setup}. It is compared against the current state-of-the-art PI$^{2}$-ES-Cov algorithm. For both cases, our approach outperforms in (1) skill success rate, (2) total accumulated reward, and (3) number of trials. 
These results show that our keyframe demonstration seeded policy search approach can be utilized to learn real-life manipulation skills with perceptual sub-goals. Within this context, our contributions are as follows:
\begin{itemize}
\item A Dynamic Bayesian Network to jointly model the action and perception.
\item Algorithmic methods to concentrate the RL exploration effort using the action and perception relationship.
\item The BO-PI² algorithm that combines black-box policy search and UCB-type return predictive exploration.
\end{itemize}

\section{RELATED WORK}

In keyframe demonstrations, the data is only collected for points that the teacher marks \cite{robot_learning_from_human_teachers}; this makes them sparse (along the time dimension) trajectories. \mycomment{They lose dynamics information which is not of primary importance for the skills we are interested in.} Keyframes can be interpreted as the consecutive sub-goals of the skill. Concentrating only on the sub-goals rather than the entire trajectory increases the generalization capacity of the methods and decreases the amount of needed data. 

Keyframes can be modeled with Hidden Markov Models (HMM) \cite{c4}. Hidden states of the HMM represent the true sub-goals, and emissions are the keyframes. Akgun et al. \cite{simultaneously_learning_actions_and_goals_from_demonstration} use two HMMs to model the changes in the object of interest and end-effector movement. They called these two models action (models the end-effector pose during the demonstration) and goal (models the perceptual representation of the object of interest). They use the action HMM to generate end-effector trajectories for skill execution, and the goal HMM to monitor this execution. 
The execution monitoring of the goal HMM can also be utilized for skill improvement. However, this gives a sparse signal for reinforcement learning, making it challenging to learn. Cem et al., convert the goal HMM to a Partially Observable Markov Reward Process (POMRP) to provide denser rewards \cite{reward_learning_from_very_few_demonstrations}.
These HMM-based approaches do not utilize the connection between the action and goal model.
In this paper, we learn a single probabilistic model to jointly represent the action and goal. We then utilize this connection to focus the reinforcement learning effort on the failed sub-goals to reduce the amount of exploration.

In recent years, model-free black-box optimization (BBO) PS methods have outperformed the traditional policy search approaches \cite{policy_improvement_methods:between_black-box_optimization_and_episodic_reinforcement_learning, a_survey_on_policy_search_algorithms_for_learning_robot_controllers_in_a_handful_of_trials}. 
Cross-Entropy Method (CEM) \cite{the_cross_entropy_method_for_fast_policy_search}, iteratively perturbs policy parameters to generate better policies. CEM sorts the perturbed parameters by their rewards and uses reward-weighted averaging to calculate the next policy using the top M parameters. Reward-weighted averaging is beneficial when the reward signal is noisy and using top M parameters decreases the effect of outliers. Matrix Adaptation Evaluation Strategy (CMA-ES) \cite{completely_derandomized_self-adaptation_in_evolution_strategies} improves CEM by storing step size and covariance matrices and allows probabilities to be any scalar reward. These increase the convergence speed. However, CMA-ES requires hand-tuned parameters, which requires additional effort for adaptation. 

Derived from stochastic optimal control, PI$^{2}$ \cite{a_generalized_path_integral_control_approach_to_reinforcement_learning} samples from a Gaussian Distribution and use probability-weighted averaging just like CMA and CMA-ES. These similarities are used in Path Integral Policy Improvement with Covariance Matrix Adaptation (PI$^{2}$-CMA) \cite{path_integral_policy_improvement_with_covariance_matrix_adaptation} approach, which has the structure of the PI$^{2}$ and covariance matrix adaptation of CEM. Authors get rid of the hand-tuned parameter of the CMA-ES and use full covariance matrix adaptation, which increases the learning speed \cite{a_survey_on_policy_search_for_robotics}.  
PI$^{2}$-CMA requires the reward signal to be dense and available for each time step. Unfortunately, this is not possible for all tasks, and in many skill executions, it is challenging to pinpoint the moment that leads to skill success.
On the other hand, CMA-ES \cite{the_cma_evolution_strategy_a_comparing_review}, is a BBO algorithm and has less strict requirements. As an alternative, PI$^{2}$-ES \cite{robot_skill_learning_from_reinforcement_learning_to_evolution_strategies}, a special case of CMA-ES, combines the PI$^{2}$ with BBO.
Finally, the PI$^{2}$-ES-Cov \cite{reward_learning_from_very_few_demonstrations} approach combines the adaptive exploration strategy of the PI$^{2}$-CMA and the black-box nature of the PI$^{2}$-ES. The authors tested PI$^{2}$-ES-Cov against the PI$^{2}$,  PI$^{2}$-CMA,  PI$^{2}$-ES on convergence speed and skill success and showed that the black-box version of the algorithm outperforms all counterparts. 


Models of the expected return can be used to improve data efficiency for learning new policies  \cite{a_survey_on_policy_search_algorithms_for_learning_robot_controllers_in_a_handful_of_trials}. These approaches refine their expected return predictions with each sample and choose the next parameters accordingly. Bayesian Optimization (BO) is a popular choice among these. BO \cite{a_tutorial_on_bayesian_optimization_of_expensive_cost_functions_with_application_to_active_user_modeling_and_hierarchical_reinforcement_learning} searches for maximum objective function value using previous samples. These methods are particularly well-fitted when sampling for an objective function that is costly to evaluate, such as in robotics. BO techniques can drastically reduce the number of trials to find a good-enough policy. Cully et al. \cite{robots_that_can_adapt_like_animals} showed that GP can be used to guide the exploration effort 
to recover the speed of a damaged robot. 
Unfortunately, BO suffers from scalability, which restricts the dimensionality of the policy. Due to this, previous work is only possible with well-chosen policy structures with small policy spaces. Furthermore, these approaches do not consider the complex, noisy reward functions, where small changes in parameter spaces can have drastic effects. 


As an alternative, methods that combine the reward-weighted averaging and exploration strategies of BBO with the reward model of the BO provide efficient and safe policy updates. One such approach Black-DROPS \cite{black_box_data_efficient_policy_search_for_robotic} uses Gaussian Process (GP) for both robot dynamics, and reward model. While this approach is data efficient, it is not scalable for learning complex and high-dimensional dynamics (e.g., when perception is involved).


We are using return modeling in episodic black box policy search, which combines the best parts of episodic BBO and BO. Our method increases data efficiency while maintaining the benefits of reward-weighted averaging, adaptive exploration and flexibility of the black-box algorithms. This is made possible by focusing the exploration with the action-perception relation. We categorize our approach as semi-BBO because of keyframe constraints in our search space.

\section{LfD FRAMEWORK}


In this section, we describe our learning from demonstration framework that learns robotic manipulation skills with perceptual (sub-)goals where dynamics is not the main component.
Our main aim is to reduce the number of trials to get successful skill models. We do this by learning a model to jointly represent actions and goals. This model is then used to focus the high-level exploration effort and combined with a reward predictive approach to improve low-level exploration.

\begin{figure}
\centering
\includegraphics[width=3in]{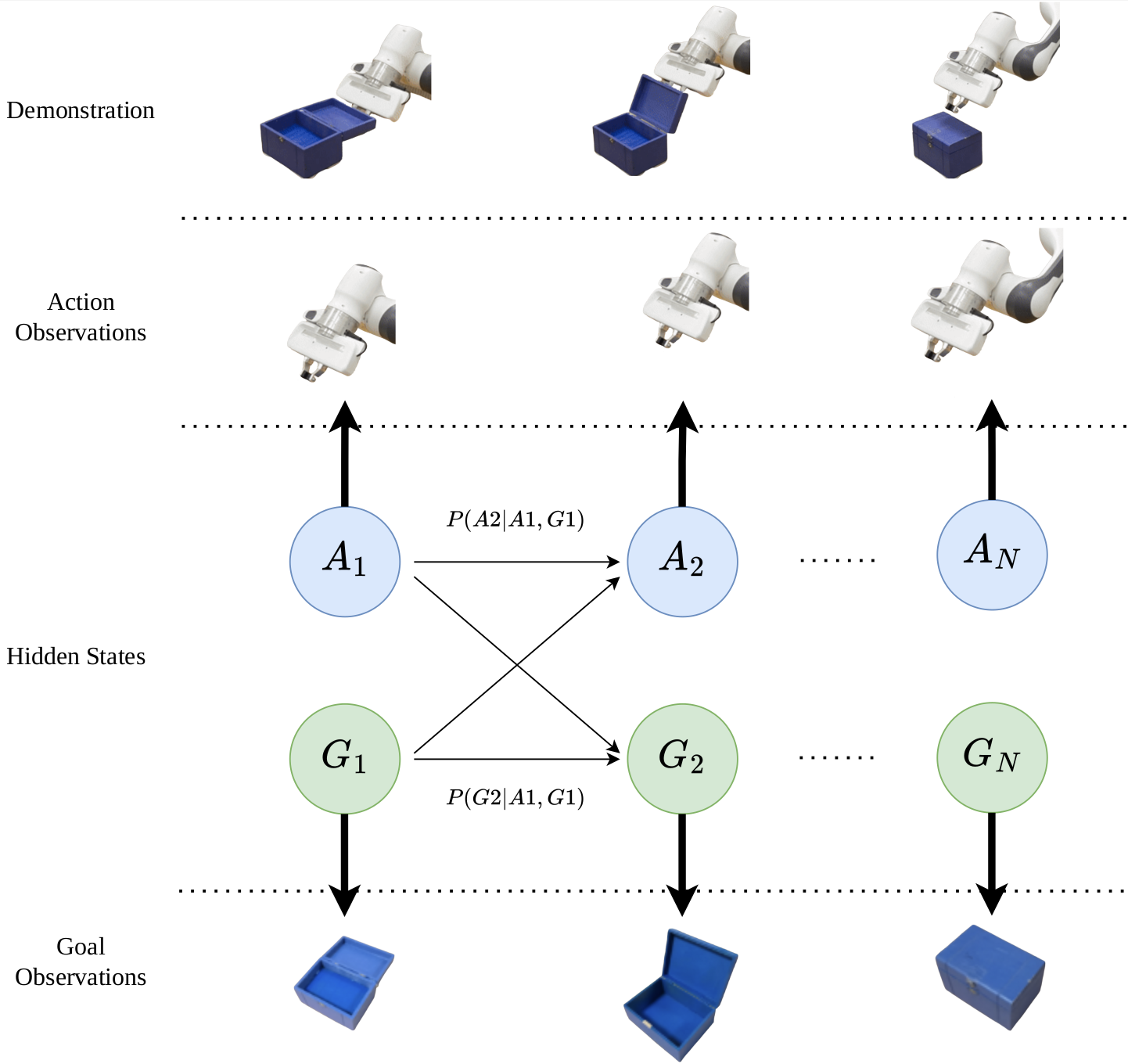} \caption[Learning action-goal demonstrations with DBN]{One DBN model is used to learn the action and goal, along with the connection between each other. Both action and goal observations are multivariate Gaussian distributions.}
\label{fig:dbn}
\end{figure}


\subsection{Action and Goal Data}
In our approach, human teachers provide keyframe data during kinesthetic demonstrations. A snapshot of an example interaction can be seen in Fig.~\ref{fig:setup}. We use the robot joint encoders to collect action data and a depth camera to collect perceptual data, which are then further processed as described next, whenever the teacher provides a keyframe.

The action observations are the pose of the end-effector with respect to the object of interest in 7D (3D position and 4D unit quaternion to represent orientation). 

The goal observations are 8D object features, obtained with the same approach as \cite{reward_learning_from_very_few_demonstrations}. We first segment the object of interest from the point cloud and calculate features in two steps. The first step is to pass the object points through a pre-trained point cloud auto-encoder \cite{learning_representations_and_generative_models_for_3d_point_clouds} to get skill agnostic features. The second step is to perform PCA on the demonstrations to get lower dimensional skill-specific features. 


\subsection{Dynamic Bayesian Network}

Our novel model, Dynamic Bayesian Network, includes the connection between the action and goal models. This new architecture, demonstrated in Fig. \ref{fig:dbn}, is very similar to HMMs \cite{c4}. We have two prior probability vectors, one for action hidden states and another for goal hidden states. Each hidden state has its own Gaussian emission parameters with properties depending on the type of hidden state. Our DBN can have different number of action and goal hidden states. 
The connection between the types of hidden states results in our DBN results in a different transition structure than HMMs. The new transitions use previous hidden states from action and goal parts rather than just action or goal. DBN requires two (one for each next goal and action state) 3D transition tensors to represent such transitions compared to HMM's 2D transition matrix. This difference still makes it possible to use EM-based parameter learning.

\subsubsection{Learning}

We learn the DBN's prior probabilities, transition probability tensors, emission model parameters, and terminal probabilities (the probability of a hidden state to be at the end of a demonstration) from demonstrations. All parameters except the transition tensors are learned using the update rules of the Baum-Welch Algorithm for HMMs. We derive custom rules for the transition tensors, but they are very similar to that of the Baum-Welch algorithm \footnote{Sharing here due to space constraints: https://bit.ly/dbn\_equations}. 

We start the E-M updates by first initializing the DBN model. We initialize the emission probabilities 
with Gaussian Mixture Models which itself is initialized ten times for best fit. The number of action hidden states is selected as the maximum number of keyframes seen during the demonstrations, and the number of goal hidden states is selected experimentally depending on the skill type. 

\subsubsection{Trajectory Generation} In order to sample trajectory to be used in RL,  we use modified version of the proposed method in \cite{simultaneously_learning_actions_and_goals_from_demonstration}. Our version samples both action and goal hidden state trajectory and generates linear path between the emission means of the action hidden states.

\subsubsection{Reward Learning} We adapt the HMM based reward learning of \cite{reward_learning_from_very_few_demonstrations} for our DBN approach. Our approach only uses goal part of the DBN to generate dense reward signals.

\section{REINFORCEMENT LEARNING}

This section explains BO-PI$^{2}$ and its novelties over previous approaches. We will first describe our algorithms for focusing on the failed sub-goals of the trajectory and then go over the novel sampling approach BO-PI$^{2}$ introduces. 

\subsection{State Selection: Focusing the High-Level Exploration}

Previous PI$^{2}$ based approaches execute the entire trajectory in a single rollout. Our approach naturally extends DBN into RL domain by breaking the trajectory into more manageable sub-goals to solve the credit attribution problem. 

We execute the initial trajectory generated by the DBN to find failed sub-goals. During the execution, we collect action and goal data at keyframe points. Then we use the Viterbi algorithm to find the most likely hidden states corresponding to the observed keyframes. We already know the expected hidden states from the DBN. Our approach compares the expected and the observed goal hidden states 
to find the failed sub-goals using Alg. \ref{alg:tuple_generation}. 

If the expected and observed goal hidden states are not the same, we consider the current sub-goal failed. In such cases, we append the current and the previous action hidden states to the improvement list (lines 6-9). The reason for adding the previous keyframe is that the current sub-goal depends on the motion between the previous and the current action state.
Note that the DBN's connection between the action and goal part allows us to use this algorithm. This approach allows us to focus on the failed sub-goals of the trajectory necessary for skill success and skip unrelated parts.


\begin{algorithm}[t]
\small
\caption{Sub-Goal Assessment }
 \begin{algorithmic}[1]
 \label{alg:tuple_generation}
 \renewcommand{\algorithmicrequire}{\textbf{Input:}}
 \renewcommand{\algorithmicensure}{\textbf{Output:}}
 \REQUIRE Expected and Observed Action/Goal Hidden States
 \ENSURE  Failed Sub-Goals
 \STATE $SubGoals$ = $\varnothing$
  \FOR {$i = 0$ to \#$Observations$}
  \IF {$ObsGoalHS[i] = ExpectedGoalHS[i]$}
  \STATE continue
  \ELSE
      \IF {$ExpectedGoalHS[i-1] \notin SubGoals$}
      \STATE $SubGoals \leftarrow ObsActionHS[i-1]$
      \ENDIF 
  \STATE  $SubGoals \leftarrow ObsActionHS[i]$
  \ENDIF
  \ENDFOR
 \RETURN $SubGoals$
 \end{algorithmic}
 \end{algorithm}

As mentioned above, the motion between two hidden states is vital in achieving the sub-goals. Thus we take pairs of consecutive action states from the output of Alg.~\ref{alg:tuple_generation} as candidates for updates and generate rollouts between the start and end states of the pair. We sample from the emission models of these states (see Sect.~\ref{sec:gp}) and generate a linear trajectory between them to execute for rollouts.

We update the emission model of a single action hidden state among the pair. We choose this approach because failure might depend on only one hidden state, and updating only one halves the parameter space. Decreasing the parameter space leads to better sample efficiency for the reward-predictive modelling that comes later. This method requires us to choose the hidden state to update among the pair. We start updating the end-state first. If the reward obtained between consecutive episodes is less than a threshold after a pre-determined number of episodes, we switch the state to update. 
We perform this until the sub-goal is satisfied or the maximum number of episodes is reached. 




\subsection{Reward Prediction: Focusing the Low-Level Exploration}
\label{sec:gp}

After selecting the action state to update, we perform Episodic RL. Our approach, BO-PI$^{2}$, utilizes Bayesian Optimization to model the expected-return and use UCB type exploration.
\ BO-PI$^{2}$ utilizes GP \cite{c1} to model the return and refine its prediction with each new execution. We use GP because it shows notable results in the literature \cite{robots_that_can_adapt_like_animals, sample_efficient_optimization_for_learning_controllers_for_bipedal_locomotion, black_box_data_efficient_policy_search_for_robotic} and gives us both the prediction and its uncertainty. 
The form of the GP can be seen below:
\begin{equation}\label{eqn:gp}
GP(x, D,  k) = \mathcal{N}(\mu(x), \sigma^{2}(x))
\end{equation}
Where x is the query point,  D is the dataset including previous observations, and k is the kernel function, which is Gaussian in this study due to its simplicity. The output of the Gaussian Process is Gaussian distribution with mean $\mu$ and standard deviation $\sigma$. 
\begin{figure}
\centering
\includegraphics[width=\columnwidth]{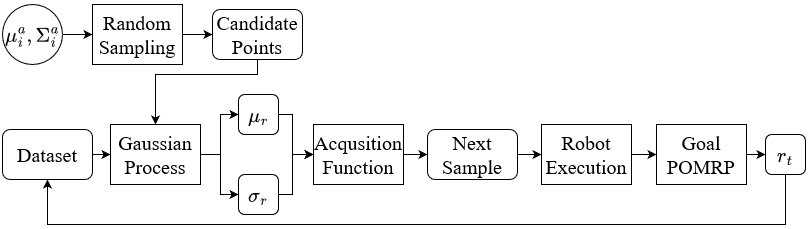}
\caption[Sampling Pipeline]{BO-PI$^{2}$ Sampling pipeline}
\label{fig:sampling_pipeline}
\end{figure}

Our GP takes 3D position information as input and outputs the expected return obtained from the learned POMRP and its variance. We exclude the 4D orientation information to work in a lower dimensional space due to the trade-off between prediction performance and data requirements.


The pipeline for GP-based sampling is depicted in Fig. \ref{fig:sampling_pipeline}. We randomly sample N candidate points during each rollout using the action hidden state emissions. GP trained with the data gathered through the previous executions is used the predict reward and uncertainty of the candidate points. Then BO-PI$^{2}$ utilizes the acquisition function to compare and select the next executed point. We use UCB \cite{an_experimental_comparison_of_bayesian_optimization_for_bipedal_locomotion} strategy as can be seen in Eqn. \ref{eqn:ucb}, to select the next sample depending on the predicted reward and the uncertainty of the candidate points. We utilize UCB because of its simplicity and success in previous studies \cite{an_experimental_comparison_of_bayesian_optimization_for_bipedal_locomotion, robots_that_can_adapt_like_animals}. UCB utilizes $\alpha$ constant to handle the trade-off between exploration and exploitation. The maximum is taken based on the N sampled points.
\begin{equation}
\label{eqn:ucb}
a_{t+1} =  \underset{x}{arg\ max}(\mu_{t}(a) + \alpha \sigma_{t}(a))
\end{equation}
At the end of each episode, BO-PI$^{2}$ uses the update rules of the PI$^{2}$-ES-Cov to update emission parameters using reward weighted averaging. The final version of BO-PI$^{2}$ can be seen in Alg. \ref{alg:bopi}, where highlighted lines show the additions of BO-PI$^{2}$ over PI$^{2}$-ES-Cov. We start by finding the sub-goals with Alg.~\ref{alg:tuple_generation} using the initial trajectory generated by the DBN (line 1). Then we run episodic RL for each sub-goal in our list. In each episode, we select the hidden state BO-PI² will focus on (line 4). After each episode (lines 3-10),  we update the model action emissions and execute the new model to check whether successful; if we succeed, we move into the next sub-goal in our list (lines 16-19).

\def\HiLi{\leavevmode\rlap{\hbox to \linewidth{\color{green!40}\leaders\hrule height .8\baselineskip depth .5ex\hfill}}}

\begin{algorithm}
\small
    \caption{BO-PI$^{2}$}
    \begin{algorithmic}[1]
    \label{alg:bopi}
        \STATE \HiLi Execute Sub-Goal Assessment
        \FOR{$SubGoal$ in $SubGoals$}
        \FOR{Episodes}
            \STATE \HiLi Update Candidate
            \FOR{Rollouts} 
                \STATE \HiLi Select the next parameter
                \STATE Execute the skill
                \STATE Calculate the return
               \STATE  \HiLi Update GP
            \ENDFOR
            \STATE Calculate the probability of rollout
            \STATE Update the parameters
            \STATE Update the exploration covariances 
            \STATE Execute the mean skill
            \STATE \HiLi Update GP
            \STATE \HiLi Calculate Success
            \STATE \HiLi \textbf{if} Success \textbf{then}
            \STATE \HiLi \phantom{for} break
            \STATE \HiLi \textbf{end if}
        \ENDFOR
        \ENDFOR
    \end{algorithmic}
\end{algorithm}



    

\section{EXPERIMENTS}

We compared BO-PI$^{2}$ and PI$^{2}$-ES-Cov methods in three metrics; skill success, cumulative reward (Normalized between 0 and 250.), and convergence speed. In order to make a reliable comparison between the algorithms, each experiment is repeated five times, and we adapt PI$^{2}$-ES-Cov to keyframes. The only distinction between BO-PI$^{2}$ and PI$^{2}$-ES-Cov is how the rollout samples are generated; compared to the BO-PI$^{2}$'s reward predictive approach, PI$^{2}$-ES-Cov uses adaptive random sampling.

Both methods employ state selection algorithm and, we did not perform experiments without these. Going over each hidden state one-by-one is trivially worse than concentrating on the failed states. Furthermore, updating all the action emissions at once would have high-variance and limit the applicability of GP due to increased dimensionality, both resulting in high sample complexities.

\subsubsection{Setup}
Our experiment setup consists of Franka Emika Panda arm and Microsoft Kinect V1 (Fig. \ref{fig:setup}). The demonstration data only consist of keyframe information. During the Reinforcement Learning we collect robot end effector and perceptual pipeline data with 10Hz.

\subsubsection{Skills}
We select three skills by considering their difficulty to demonstrate on our robotic arm. Both expert and naive users gave demonstrations of three different skills; \emph{open the box}: robot opens a box, \emph{close the box}: robot closes a box, and \emph{open the drawer}: robot pulls to open a wooden drawer. Objects for each skill can be seen in Tab.~\ref{fig:objects}.

\begin{table}
\renewcommand{\arraystretch}{1.3}
\caption{Object used in Experiments}
\label{fig:objects}
\centering
\begin{threeparttable}
\begin{tabular}{c||c||c}
\hline
\bfseries Name & \bfseries Close & \bfseries Open   \\
\hline\hline
Box1   &  \includegraphics[width=0.3in]{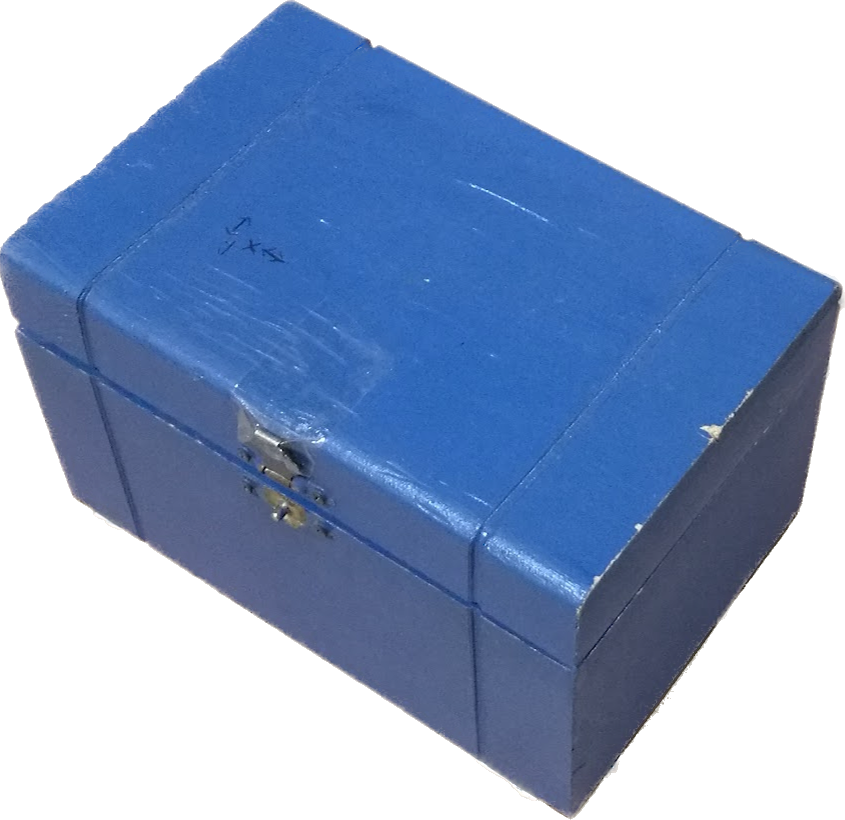} &  \includegraphics[width=0.3in]{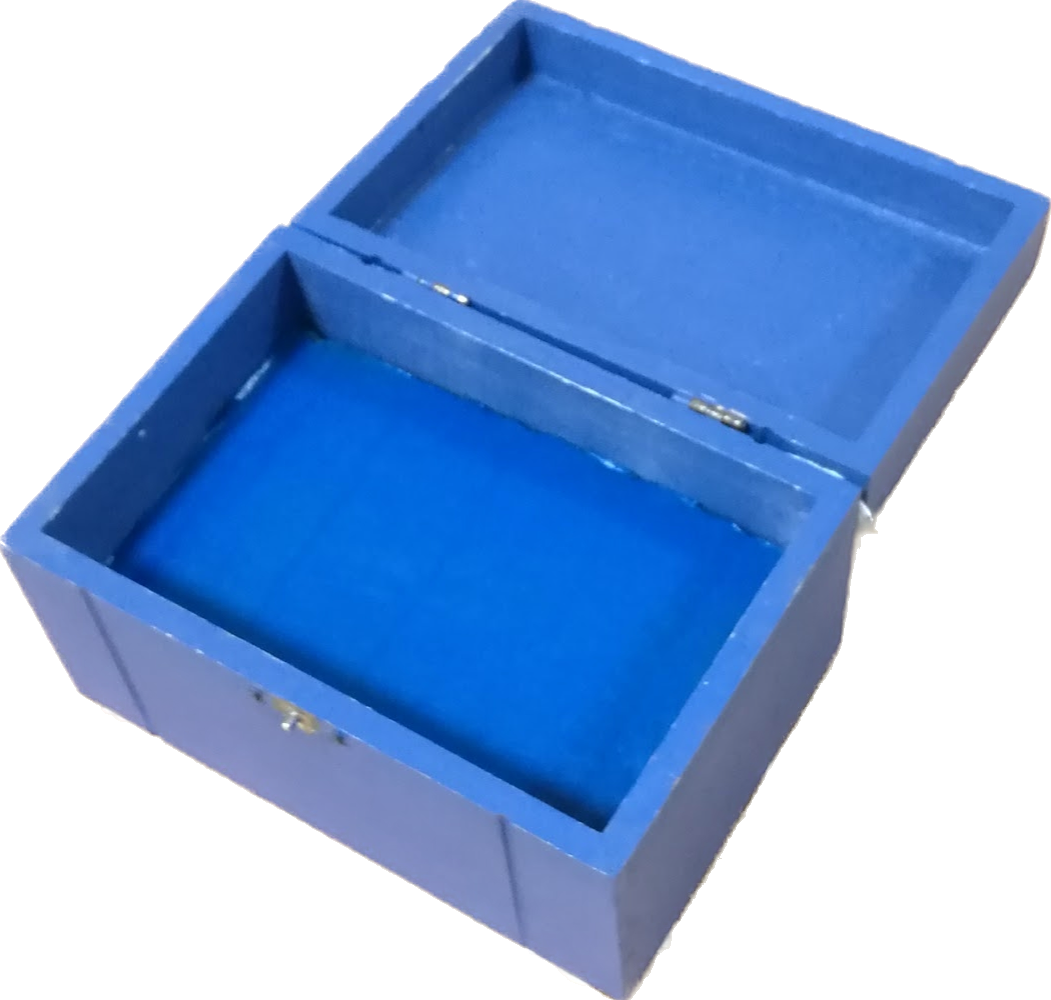} \\Box2   &  \includegraphics[width=0.3in]{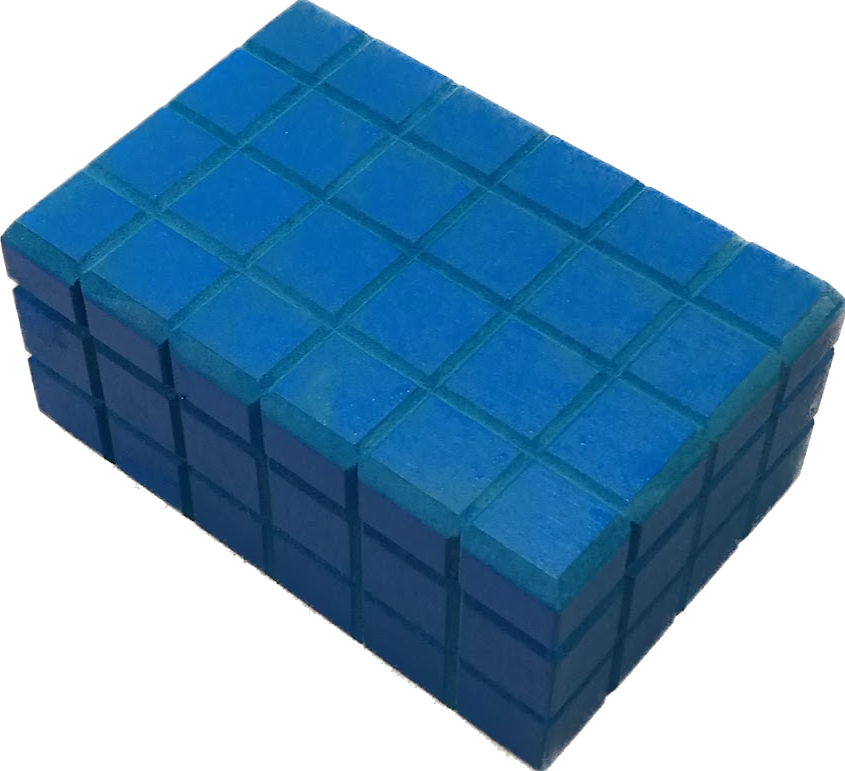} &  \includegraphics[width=0.3in]{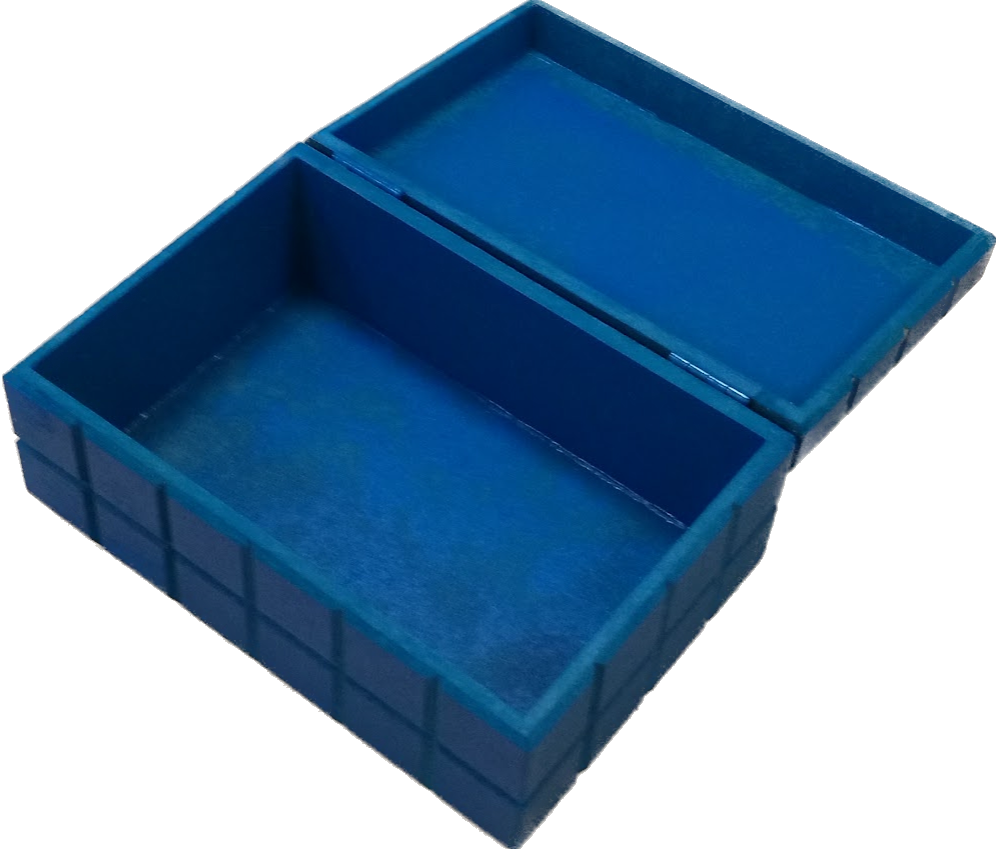} \\
Drawer &  \includegraphics[width=0.3in]{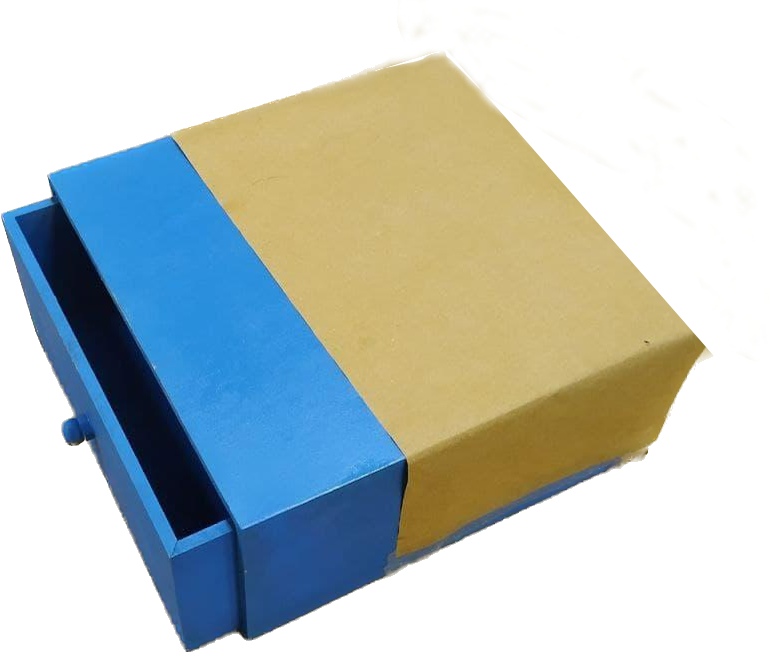} & \includegraphics[width=0.3in]{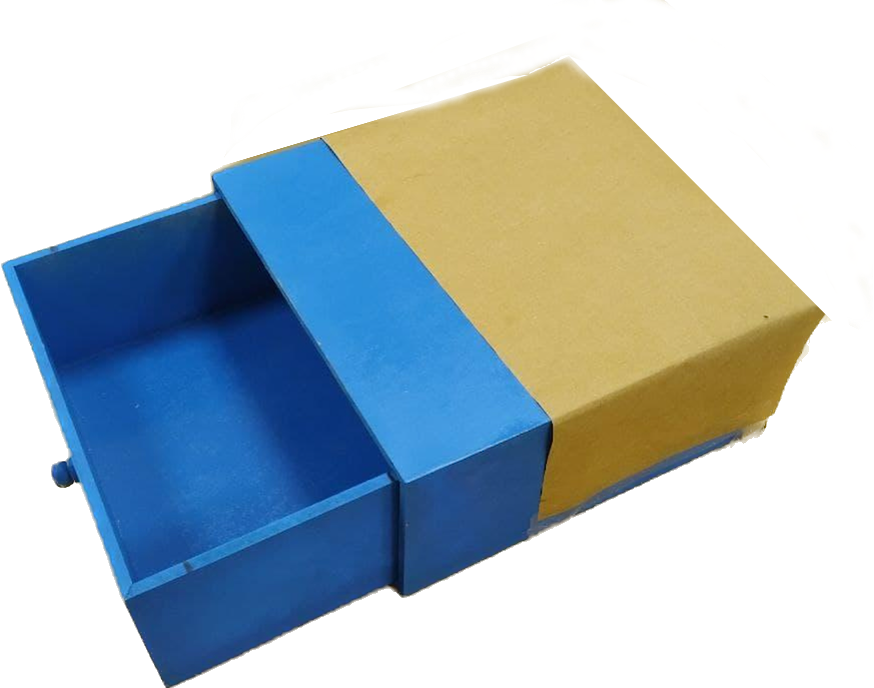}\\
\end{tabular}
\end{threeparttable}
\end{table}



\subsection{Expert Demonstrations}

We tested four object-skill combinations with expert user demonstrations. Initial DBN models learned from the demonstrations were successful. To start Reinforcement Learning from a fail state, we manually perturb the means of the hidden states. The perturbations 
can be seen in Tab.~\ref{table:perturbations}.

\begin{table}
\renewcommand{\arraystretch}{1.3}
\caption[Random mean perturbations for each hidden state]{Random mean perturbations for each hidden state$^{*}$}
\label{table:perturbations}
\centering
\begin{threeparttable}
\begin{tabular}{c||c||c||c||c||c}
\hline
\bfseries Skill & \bfseries Object Type & \bfseries First & \bfseries Second & \bfseries Third & \bfseries Fourth \\
\hline\hline
Close & Box 1 & -    & -    & 9    & 6    \\
Close & Box 2 & -    & -    & 10.2 & 14   \\
Open &  Box 2 & -    & 5    & -    & 10.4 \\
Open & Drawer & -    & -    & 9.8  & -    \\
\hline
\end{tabular}
\begin{tablenotes}
\item[*] Units in centimeter.
\end{tablenotes}
\end{threeparttable}
\end{table}

\begin{table*}
\renewcommand{\arraystretch}{1.3}
\caption{Expert Demonstration Results averaged over 5 experiments.}
\label{table:expert_results}
\centering
\begin{threeparttable}
\begin{tabular}{c||c||c||c||c||c||c||c}
\hline 
\textbf{Skill} & \textbf{Object} & \multicolumn{2}{c||}{\textbf{Skill Success}} & \multicolumn{2}{c||}{\textbf{Return}} & \multicolumn{2}{c}{\textbf{Term. Time\textdagger}} \\\cline{3-8}
      &        & BO-PI$^{2}$           & PI$^{2}$-ES-Cov & BO-PI$^{2}$         & PI$^{2}$-ES-Cov     & BO-PI$^{2}$        & PI$^{2}$-ES-Cov  \\
\hline\hline
Close & Box 1  & \textbf{100}\%  & 40\%      &  \textbf{213} & 76            & \textbf{3.2} & 6.4        \\
Close & Box 2  & \textbf{100}\%  & 20\%      &  \textbf{198} & 11            & \textbf{4.6} & 5.8        \\
Open  & Box 2  & \textbf{100}\%  & 20\%      &  \textbf{146} & 25            & \textbf{5.6} & 5.8        \\
Open  & Drawer & 80\%            & 80\%      &  171          & \textbf{174}  & \textbf{4.6} & 4.8        \\
Avg.  & -      & \textbf{95}\%   & 40\%      &  \textbf{182} & 71.5          & \textbf{4.5} & 5.7               
\end{tabular}
\begin{tablenotes}
\item[\textdagger] Units in episode.
\end{tablenotes}
\end{threeparttable}
\end{table*}

\begin{table*}
\renewcommand{\arraystretch}{1.3}
\caption{HRI Results averaged over 5 experiments}
\label{table:hri_results}
\centering
\begin{threeparttable}

\begin{tabular}{c||c||c||c||c||c||c||c||c||c}
\hline
\textbf{Skill} & \textbf{Object} & \multicolumn{2}{c||}{\textbf{Skill Success}} & \multicolumn{2}{c||}{\textbf{Return}} & \multicolumn{2}{c||}{\textbf{Term. Time}}  & \multicolumn{2}{c}{\textbf{Euc./Ang. Disp.\textdagger}} \\\cline{3-10}
      &        & BO-PI$^{2}$           & PI$^{2}$-ES-Cov & BO-PI$^{2}$         & PI$^{2}$-ES-Cov     & BO-PI$^{2}$        & PI$^{2}$-ES-Cov & BO-PI$^{2}$   & PI$^{2}$-ES-Cov \\
\hline\hline
Close & Box 2  & 100\%            & 100\%   & \textbf{198} & 108          & 2.8          & \textbf{2.0} &  9.42/4.50   & 5.86/2.70 \\
Close & Box 2  & \textbf{80\%}    & 20\%    & \textbf{110} & 56           & \textbf{4.0} & 5.6          &  15.74/12.59 & 16.40/12.58 \\
Open  & Box 2  & \textbf{100\%}   & 60\%    & \textbf{195} & 109          & \textbf{4.8} & 5.6          &  3.87/3.58   & 1.92/2.63 \\
Open  & Box 2  & \textbf{100\%}   & 80\%    & \textbf{170} & 112          & \textbf{2.0} & 2.8          &  8.74/2.94   & 5.33/4.08 \\
Open  & Drawer & 60\%             & 60\%    & 127          & \textbf{133} & \textbf{4.6} & 5.6          &  7.18/5.03   & 6.83/6.18 \\
Open  & Drawer & \textbf{80}\%    & 60\%    & \textbf{164} & 129          & \textbf{4.0} & 4.2          &  7.79/7.07   & 5.76/5.37 \\
Avg.  & -      & \textbf{86.67}\% & 63.33\% & \textbf{160} & 107          & \textbf{3.7} & 4.3          &  8.79/5.95   & 7.01/5.59 \\              
\end{tabular}
\begin{tablenotes}
\item[\textdagger] Units in cm and degree.
\end{tablenotes}
\end{threeparttable}
\end{table*}

The overall results are shown in Tab.~\ref{table:expert_results}. BO-PI$^{2}$ outperforms PI$^{2}$-ES-Cov for almost all the skills. On average, BO-PI$^{2}$ (1) reaches 95\% skill success vs. 40\%, (2) accumulates 2.5 more rewards, and (3) terminates faster with 4.5 number of episodes vs. 5.7. These results imply that BO-PI$^{2}$ reaches higher success rates faster. This is more prevalent for larger perturbations. We argue that using a return-predictive exploration approach is better at finding high-reward regions. 



Success rate and average termination time results indicate that our state selection algorithms perform better than applying episodic RL on each hidden state of the trajectory. Without these, we would have spent unnecessary effort on non-perturbed states that do not impact sub-goal success. 

\begin{figure*}[h]
     \centering
     \begin{subfigure}[b]{0.3\linewidth}
    \centering
    \includegraphics[width=\linewidth]{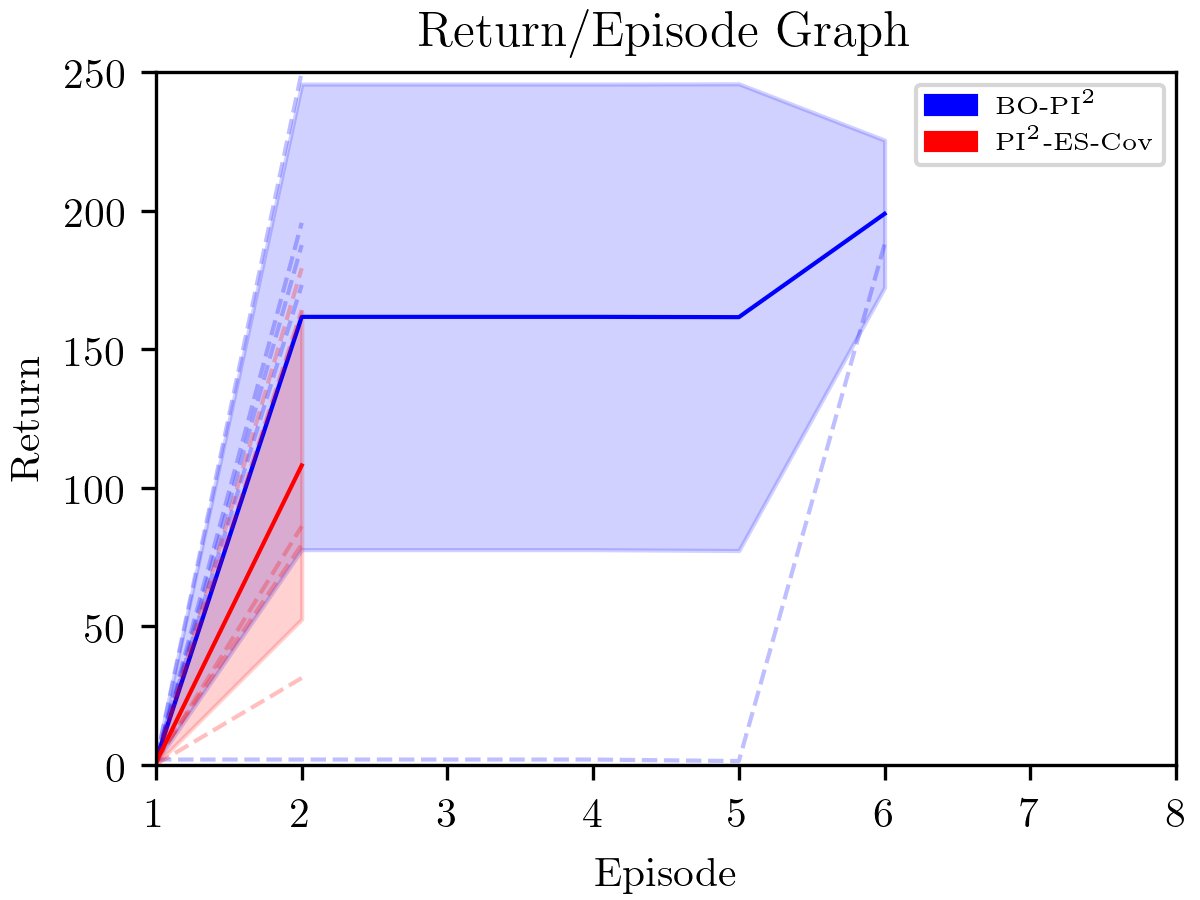}
    \caption[HRI close the box reward graph for participant 2]{Close-Box (1)}
    \label{fig:stacked:first}
     \end{subfigure}
     \hfill
     \begin{subfigure}[b]{0.3\linewidth}
         \centering
\includegraphics[width=\linewidth]{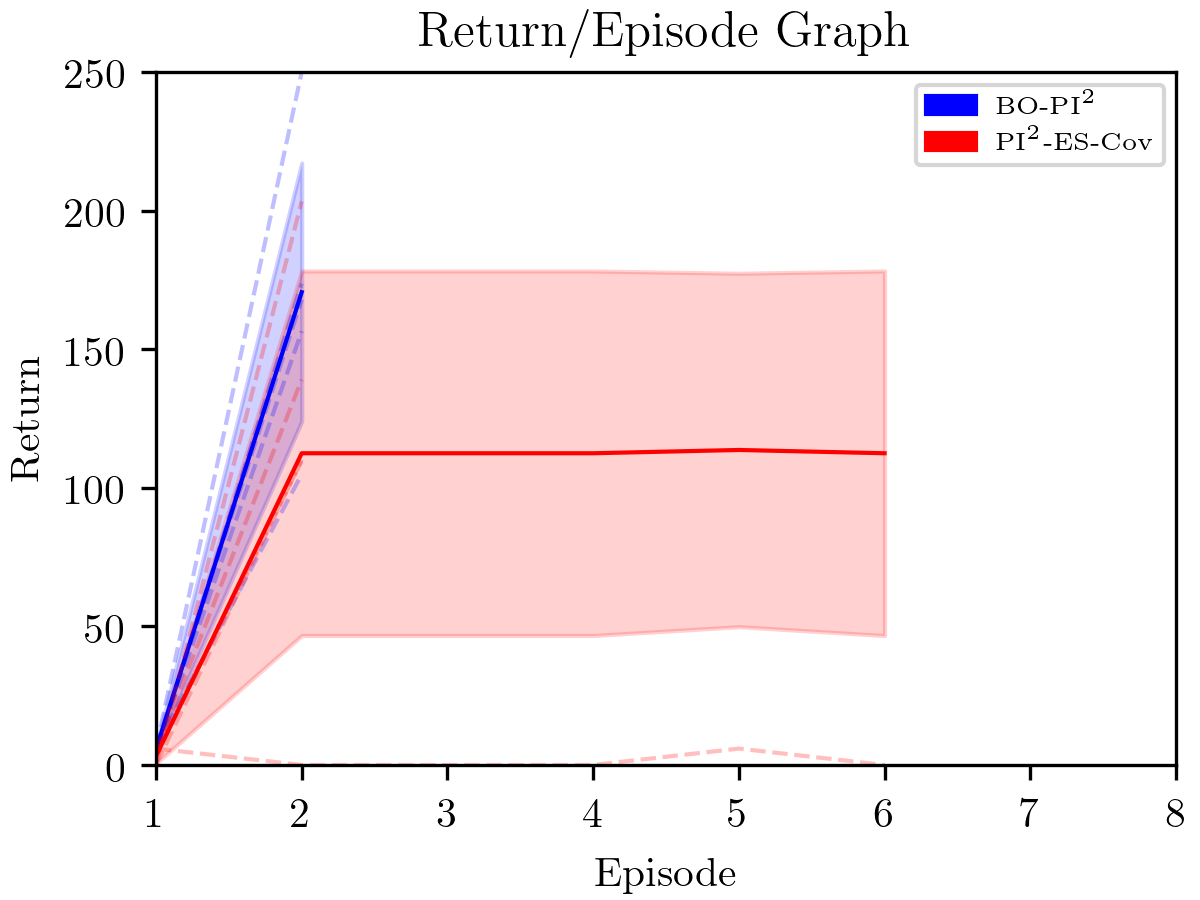}
\caption[HRI open the box reward graph for participant 7]{Open-Box (4)}
\label{fig:stacked:fourth}
     \end{subfigure}
     \hfill
     \begin{subfigure}[b]{0.3\linewidth}
         \centering
\includegraphics[width=\linewidth]{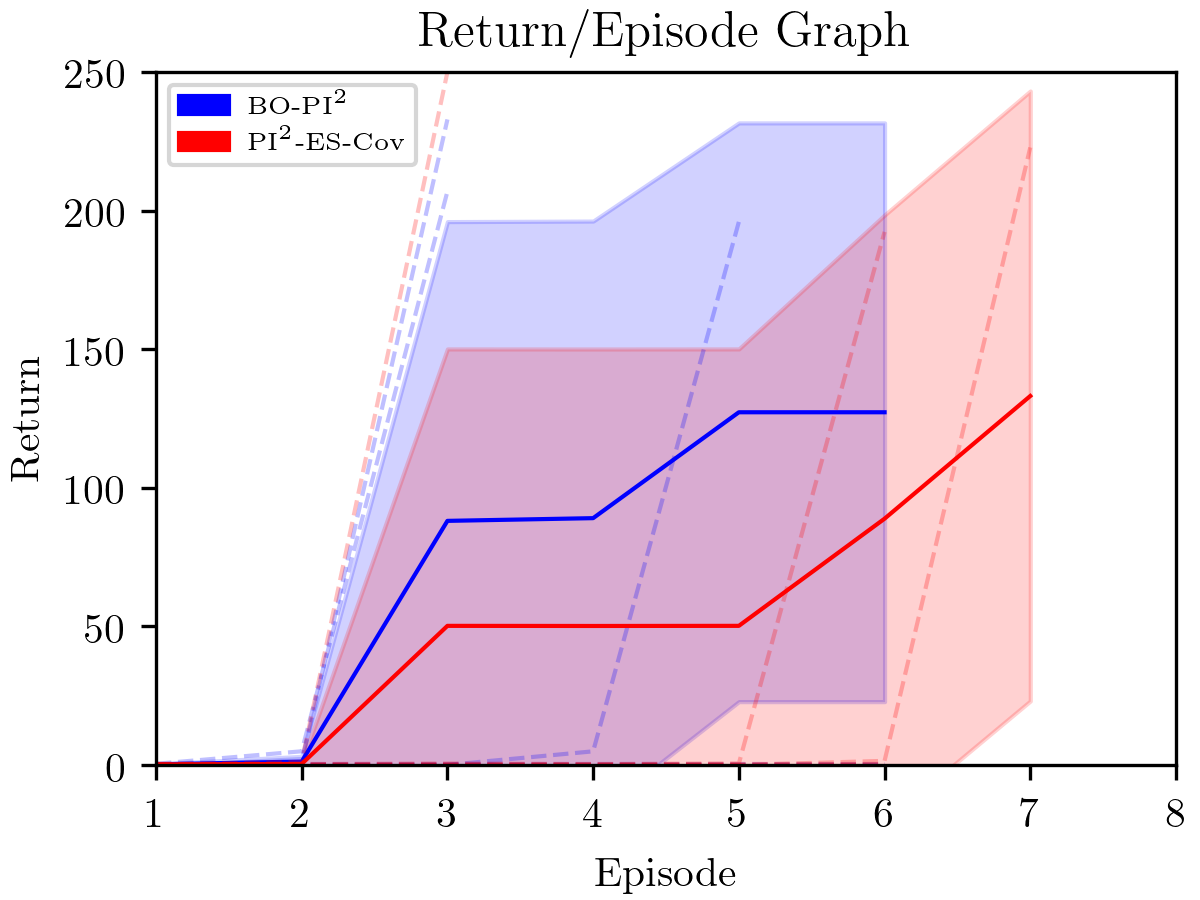}
\caption[HRI open the box reward graph for participant 4]{Open-Drawer (5)}
\label{fig:stacked:fifth}
     \end{subfigure}
        \caption{Selected return graphs for the HRI evaluations. The numbers correspond to the row number of Tab.~\ref{table:hri_results}. Transparent regions show one standart deviation. Dashed lines show individual experiments and the solid lines show the running average.}
        \label{fig:hri_returns}
\end{figure*}

\subsection{HRI experiments}
We further validate our results with an HRI study. We collect data from 8 (4 female, 4 male) naive users aged 22-28 who do not have any experience teaching a robot. Participants are selected from the campus community.

We divide the data collection protocol into 4 phases; First, the experimenter asks the user to sign the consent form and gives a brief verbal description of the study. After this part, the participant is asked to complete two simple toy problems; (1) setting the robot into specific configurations and (2) teaching the robotic arm to move between two points. During the demonstration, the user gives verbal commands to mark keyframes. Finally, the experimenter asks the user to demonstrate three skills; the order of skills is counterbalanced to minimize the effects of familiarity. Each user gives six demonstrations for each skill and is allowed to see the learned DBN model after four demonstrations. 

We pick 6 cases where the learned skills were failures and perform 5 experiments for each. The overall results are shown in Tab.~\ref{table:hri_results}. The results show that BO-PI² is either better than PI$^{2}$-ES-Cov, or they perform about the same. On average, BO-PI$^{2}$ (1) reaches 86.67\% skill success vs. 63.33\%, (2) accumulates 1.5 more rewards, and (3) terminates faster with 3.7 number of episodes vs. 4.3. Our results show that BO-PI$^{2}$ shows promising results compared against the PI$^{2}$-ES-Cov, regardless of the demonstration type, and reward predictive exploration strategies are beneficial to improve RL performance and to reduce the number of trials.

The results are not as pronounced as the perturbed expert case. We look at the number of updated action-state pairs and the amount of change in the emission space for the successfully learned skills, as a proxy to the "error" in demonstrations. For all the HRI cases, only one pair needed to be updated with less than 9 cm position change, which is less than the manual perturbations. Furthermore, the gap between the performance of BO-PI² and PI$^{2}$-ES-Cov increases with the needed amount of displacement, which, along with the expert results, imply that BO-PI² is better for harder tasks.

The Fig.~\ref{fig:hri_returns} shows the episode returns for a subset of the HRI evaluations. We present one for each skill and picked the ones with similar success rates. In Fig.~\ref{fig:stacked:first}, both of the methods reached 100\% success. One experiment of BO-PI$^{2}$ needed 6 episodes but overall reached higher returns. In Fig.~\ref{fig:stacked:fourth}, PI$^{2}$-ES-Cov could not reach success in one episode. In both of these cases, BO-PI$^{2}$ reached higher returns. In Fig.~\ref{fig:stacked:fifth}, both methods failed twice but BO-PI$^{2}$ was faster for the successful cases. These show the overall behavior of BO-PI$^{2}$ being faster and/or reaching higher returns. The differences are more apparent in the cases that were not displayed.

\section{CONCLUSION}

In this work, we introduced a keyframe demonstration seeded policy search approach called Bayesian-Optimized PI$^2$ (BO-PI$^2$). We develop a novel Dynamic Bayesian Network with two hidden states to jointly model robot motion and perceptual object features. We use the action part to execute the learned skill and treat the perceptual part as its sub-goals. We use the action-perception relation to figure out the parts of the action that lead to sub-goal failure and focus our reinforcement learning effort. We also use a return-predictive approach and UCB type exploration to further reduce this effort. Our results on both expert and naive user demonstration show that BO-PI$^2$ outperforms the state-of-the-art baseline on skill success, execution time and cumulative rewards. These results imply that BO-PI$^2$ can be used to endow robot arms with real-life manipulation skills with perceptual sub-goals.

Our work also has some limitations and potential avenues for improvement. Our DBN can learn a branching policy (i.e. different ways to execute the skill), but we only concentrate on the most likely path. Another loop over the episodes would handle this, but it may also lead to worse termination times. This is an interesting avenue to further explore. Furthermore
our approach is also not suitable for skills when the dynamic components are important. A next step would be to extend BO-PI$^2$ to work with trajectories.






\end{document}